\documentclass[letterpaper, 10 pt, conference]{ieeeconf}

\usepackage{geometry}
\IEEEoverridecommandlockouts 
\usepackage{amsmath}
\usepackage{graphicx}
\usepackage{subfigure}
\usepackage[caption=false]{subfig}
\usepackage{multirow}
\usepackage{stfloats}
\usepackage{siunitx}
\usepackage{booktabs}
\usepackage{longtable}
\usepackage{cite}
\usepackage[table]{xcolor}
\usepackage[normalem]{ulem}
\useunder{\uline}{\ul}{}
\usepackage{comment}
\usepackage{makecell}

\title{\LARGE \bf
GraspLook: a VR-based Telemanipulation System with R-CNN-driven Augmentation of Virtual Environment
}

\author{
Polina Ponomareva, 
Daria Trinitatova, 
Aleksey Fedoseev, 
Ivan Kalinov, 
and Dzmitry Tsetserukou
\thanks{The authors are with the Space Center, Skolkovo Institute of Science and Technology (Skoltech), 121205 Bolshoy Boulevard 30, bld. 1, Moscow, Russia. {\tt\small \{polina.ponomareva, daria.trinitatova, aleksey.fedoseev, i.kalinov, d.tsetserukou\}@skoltech.ru\ }}%
}

\begin{document}

\maketitle
\thispagestyle{empty}
\pagestyle{empty}

\begin{abstract}

The teleoperation of robotic systems in medical applications requires stable and convenient visual feedback for the operator. The most accessible approach to delivering visual information from the remote area is using cameras to transmit a video stream from the environment. However, such systems are sensitive to the camera resolution, limited viewpoints, and cluttered environment bringing additional mental demands to the human operator. The paper proposes a novel system of teleoperation based on an augmented virtual environment (VE). The region-based convolutional neural network (R-CNN) is applied to detect the laboratory instrument and estimate its position in the remote environment to display further its digital twin in the VE, which is necessary for dexterous telemanipulation.
The experimental results revealed that the developed system allows users to operate the robot smoother, which leads to a decrease in task execution time when manipulating test tubes. In addition, the participants evaluated the developed system as less mentally demanding (by 11\%) and requiring less effort (by 16\%) to accomplish the task than the camera-based teleoperation approach and highly assessed their performance in the augmented VE.
The proposed technology can be potentially applied for conducting laboratory tests in remote areas when operating with infectious and poisonous reagents.

\end{abstract}

\section{Introduction}

The use of telemanipulation in medicine \cite{Mehrdad_2021} and Industry 4.0 \cite{gao2020industry} has been the subject of interest for many research groups over the past decade due to its high impact on the precision and safety of critical operations, e.g., inspection, assembling, surgery, manipulation with dangerous reagents, etc. These operations, however, are often linked to dynamic environment changes, which require providing constant visual feedback to the operator. Visual information from the environment is the most natural for the operator since it does not require any training and additional equipment aside from the regular camera installation. 
A lot of research works have been done to implement visual feedback from RGB cameras for such purposes. Acemoglu et al. \cite{Acemoglu_2020} proposed a telesurgery robotic system, where visual feedback is obtained from the cameras with low latency due to the 5G implementation. Zakharkin et al. \cite{Zakharkin_2020} proposed a teleoperation system, which utilizes a Zoom application camera not only as a tool of feedback but also as an active control interface. 
However, teleoperation through visual feedback limited to the camera view only shows high efficiency when the robot is located at the center of the image without any cluttered environment.

The integration of VE into teleoperation control loop is considered as a way to increase intuitiveness and awareness of human-robot interaction (HRI) with augmented visual feedback to the operator \cite{Wonsick_2020}.
Kalinov et al. \cite{kalinov2021warevr} developed a virtual reality (VR) interface for supervision of autonomous robotic system aimed at warehouse stocktaking, which enhanced the operator's ability via a preliminary simulation of the task in a virtual scene.

In the research works of Lima et al. \cite{Tonel2018}, the kinesthetic feedback from Geomagic Touch X haptic device was combined with multiple visual channels from the real robot and its real-time simulation. Lipton et al. \cite{Lipton_2018} introduced a virtual reality-based teleoperation interface for a bimanual robot, propagating view from several cameras into one virtual scene. Ni et al. \cite{ni2017haptic} developed an interface for programming welding robots using desktop haptic device and augmented reality.

\begin{figure}[!t]
\begin{center}
\subfigure[Operator with HMD during teleoperation process.]{
\includegraphics[width=0.97\linewidth]{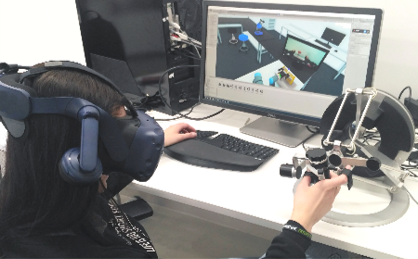}}
\subfigure[Isometric view of the virtual scene.]{
\includegraphics[width=0.47\linewidth]{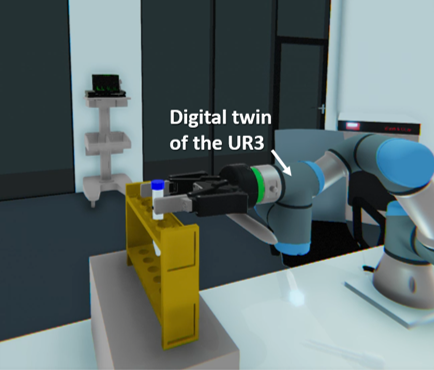}}
\subfigure[Robotic arm during teleoperation process.]{
\includegraphics[width=0.47\linewidth]{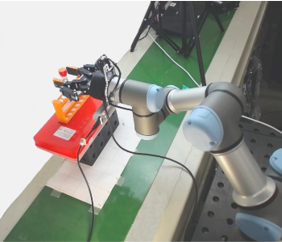}}
\caption{GraspLook telemanipulation system.}\label{SystemOverview}
\end{center}
\vspace{-2 em}
\end{figure}
Among the implementations of an augmented visual environment, the 3D reconstruction shows high potential. It allows the operator to interact with both pre-defined objects of the deterministic environment and objects, which location and properties might change during the teleoperation scenario.
Denninger et al. \cite{Denninger_2020} introduced a novel approach to infer volumetric reconstructions from a single viewport, with occluded region reconstruction by synthetically trained deep neural network.
Omarali et al. \cite{Omarali2020} proposed a VR-based teleoperation system with dynamic field of view to control the robot. As a visual channel, the Intel RealSense d415i and Kinect2 depth cameras were for efficient and accurate capture of objects and displaying them in the virtual scene. Kohn et al. \cite{Kohn2018} proposed a real-time teleoperation system using depth cameras to reconstruct the environment of the robot in VR. For the reconstruction of objects in VR, object segmentation based on point cloud filtering algorithms was used. 
The most effective neural model in recent years was proposed by Jeong et al. \cite{Jeong2019}. It was applied to determine the position and distance to an object using a stereo camera for providing these data to capture the object by the robot. 
Sumigray et al. \cite{Sumigray2021} proposed a teleoperation system that uses VR with the reconstruction of the robot real environment using a multi-camera system. The multi-camera system comprised the Insta 360 Pro 2 camera and depth cameras located on the sides of the robot. The developed Matryoshka algorithm was used to determine the location of objects in the scene in real-time. The operator could control the position and orientation of the end effector operation using VR joysticks, using visual feedback in the Unity VR application.
 
In this work, we propose a novel telemanipulation system for the robot through VR. Objects for manipulation detected using a developed computer vision (CV) system are displayed in the VE in real-time, allowing the operator to perform precise grasping. 

\section{System Overview}\label{Overview}

\begin{figure}[!h]
\includegraphics[width=0.95\linewidth]{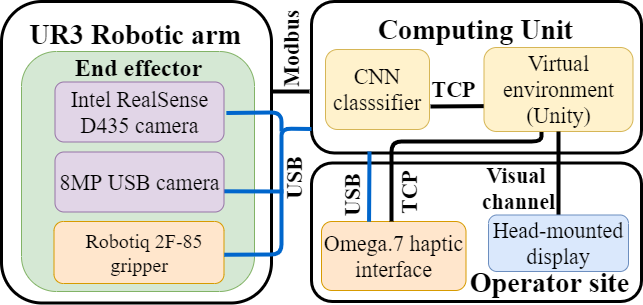}
\caption{The overall architecture of the developed system. The operator controls the collaborative robot using Omega.7 haptic device. The object for manipulation is reconstructed by a developed R-CNN-based system using Intel RealSense depth camera.}
\label{fig:system}
\vspace{-1em}
\end{figure}

The overall scheme of the developed system is presented in Fig. \ref{fig:system}. The proposed system consists of a robotic manipulator UR3 with 6 degrees of freedom equipped with a 2-finger gripper (Robotiq 2F-85), a digital 8-megapixel camera with a 75-degree field of view, and Intel RealSense D435 RGB-D camera on the end effector. The operator controls the robot using the Omega.7 desktop haptic device. As a visual feedback channel for the operator, a virtual environment is created with the Unity Engine. It includes a digital twin of the robot, a real-time video stream from the 8 MP camera located on the end effector, and a digital twin of a laboratory instrument which position is estimated with a developed CV system using Intel RealSense depth camera.


\subsection{Control system of the robot}
Positional control of the robot movement is carried out by moving the handle of the haptic device. Three translational axes of the robot EE correspond to three translational axes of the haptic device, while the EE rotations are fixed. The workspace of the haptic device (R = 160 mm) and the robot (R = 500 mm) differs from each other. Therefore, the developed control framework performs the workspace scaling from haptic device to robot by either x1, x2, x3, x4, or x5 scale factor. To increase the accuracy and speed of the robot manipulation, the movement of robot EE along one or two translational axes could be locked.
The control signal of three positional coordinates for the EE of the robot and the gripper state command are generated from the haptic device and transmitted to the control framework via sockets, thus minimizing data transmission delays. 
Since both the working area and the coordinate system of the haptic device and the robotic arm are different, the transformation and scaling of the input coordinates are performed at the first calculating stage. After that, the digital twin solves the inverse kinematics task in the Cartesian coordinate space of Unity3D. Since the UR robots are designed with a non-spherical joint configuration, a closed-form kinematic solution is chosen for real-time control with the eight possible joint configurations for one EE position. The solution is selected with the weighted Least Squares method to minimize the difference with the previous position of the robot and perform a smooth shift across all its joints.

\subsection{Digital twin}
\label{VE}
The approaches using a video stream from a remote site to obtain visual feedback during teleoperation have a number of limitations. For example, the obtained visual channel could suffer from optical distortion and camera occlusion, preventing the operator's access to the overall state of the system. To eliminate these disadvantages, we propose to use a digital twin of the robot and augment the virtual environment with 3d models of objects for manipulation in real-time mode. The digital twin and communication interface of the UR3 robot have been designed for the virtual environment in the Unity Engine \cite{Fedoseev_2019}. The digital twin provides information about the position of the real robot with a frequency of 50 Hz. The key feature of the VR interface is that it allows to represent the current state of the robot from different perspectives without any complex camera setup (Fig. 1(b)). Visual information about the gripper state supports the control of the object manipulation.

\subsection{Virtual environment augmentation}

To achieve user awareness of the manipulation process, we propose to update the state of the robot and operated object with the real-time data of digital twin and CNN-based object pose estimation. The instance segmentation of the laboratory instruments was accomplished by using trained R-CNN on a collected dataset. We use color images from RGB-D RealSense camera, which is attached to the robot EE for object detection. We have two approaches to extract the coordinates of the objects using the depth map related to the color frame: from the bounding box where the object is detected and from the predicted segmented mask. Depth map pixels for the bounding box are filtered to prevent the robot from exceeding the workspace boundaries. 
At this work, we assume that all objects are placed vertically. Therefore, it is possible to compute the average distance to the object from the filtered bounding box and segmented mask. The coordinates of the object center are calculated from the obtained point cloud. The laboratory equipment models were pre-designed in CAD software and exported to the Unity prefabs.
Thus, once the coordinates of the object center are obtained in the RealSense coordinate system and transformed into the digital twin coordinate system, the object is displayed in the VE.

\section{R-CNN-based Approach for Virtual Environment Augmentation}
\subsection{Dataset Collection}
\begin{figure}[h]
\centering
\includegraphics[width=0.45\textwidth]{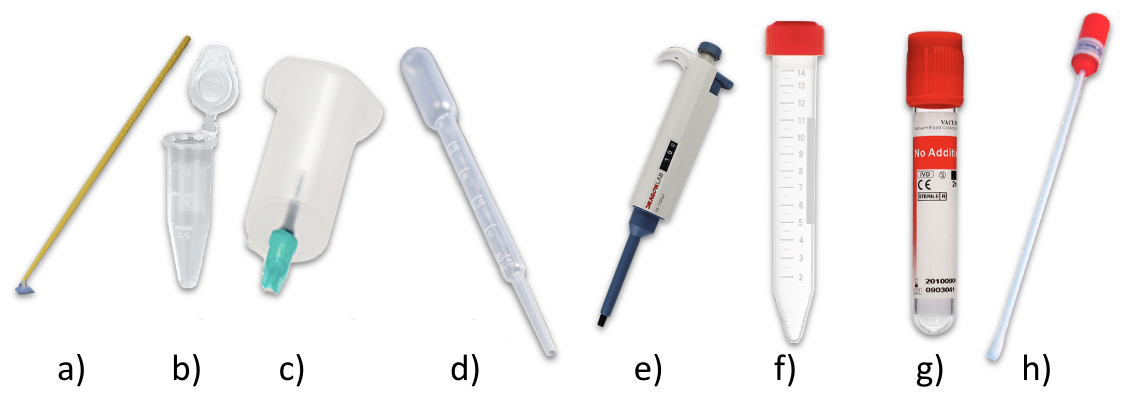}
\vspace{-1em}
\caption{Object classes of laboratory instruments in the dataset: a) cell scrapper, b) micro test tube, c) needle holder, d) Pasteur pipette, e) pipettor, f) centrifuge test tube, g) vacuum test tube, h) swab.}
\label{fig:datasetClasses}
\end{figure}

To reconstruct laboratory equipment in the VE, it is necessary to teach the system to recognize various objects. 
We created a synthetic dataset consisting of eight instruments that are most commonly used in the laboratory: scrapper, micro test tube, needle holder, Pasteur pipette, pipettor, centrifuge test tube, vacuum test tube, and swab (Fig. \ref{fig:datasetClasses}). 

Creating a dataset consisted of 3 main stages: collecting images of objects for each class, superimposing an image of the class on the background image, and annotating the image for each superimposed object. To collect images for each class, we found 30 images per class on the Internet and then deleted the background for each image. Hereafter, the images of the objects were placed on a transparent background, where the alpha channel was a mask of the object. To create a single image for the dataset, we used one random image from each class, overlaying them randomly in a random place on the background image.

\begin{figure}[!t]
\begin{center}
\subfigure[Superimposed image.]{
\includegraphics[width=0.46\linewidth]{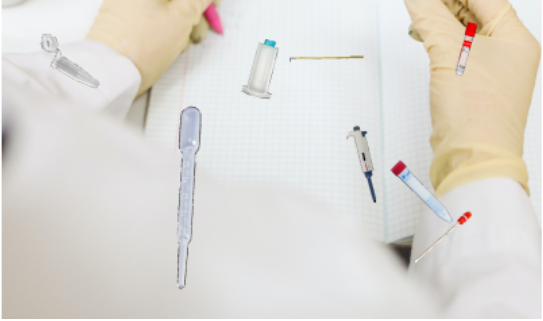}}
\subfigure[Segmented object masks for the corresponding image.]{
\includegraphics[width=0.47\linewidth]{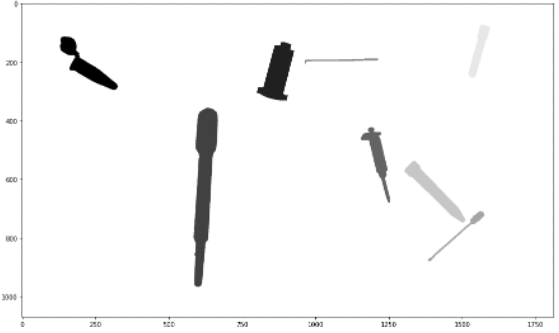}}
\caption{Sample image from the synthetic dataset.}\label{fig:dataset_seg}
\end{center}
\vspace{-2 em}
\end{figure}

When overlaying object images, we used additional augmentations: rotation, resizing, and reflection of objects. To annotate an image in accordance with the Common Objects in Context (COCO) format \cite{microsoftCOCO}, it is necessary to extract the coordinates of the mask borders for each object. Thus, at the stage of forming a set of image data, the original masks of objects are determined (Fig. \ref{fig:dataset_seg}). After superimposing them on the general background, it is sufficient to subtract the masks from each other in case of overlap, depending on the order of objects superimposed on the background image. To create the dataset, we used ten different background images with medical laboratories, which were randomly selected to generate the image of the dataset. As the result, the dataset consisting of 8000 images was created, where 6000 images were used for training and 2000 images for testing (every image contains eight object instances, one for each object class).

\subsection{Mask R-CNN training}

We used the state-of-the-art neural network architecture Mask R-CNN with the backbone ResNet-101-FPN \cite{he2017mask} in our system. We trained this model on Google Colab resource for 10k iterations with a learning rate of 0.002, and four images per batch. The evaluation of the model is reported in the standard pixel COCO metrics: AP (average precision at IoU threshold), AP$_{50}$, AP$_{75}$, and AP$_{M}$ (AP at different scales). 
It is worth noting that this class of metrics involves pixel-by-pixel comparison and gives not entirely accurate data on the number of objects. Therefore, we used the second class of metrics with quantitative measures: 
correctly recognized objects (True positive - TP) with intersection more or equal to 50\% of source object, mistakenly detected (False Positive - FP) with intersection less than 50\%, and undetected (False Negative - FN).
The best results for all metrics were obtained for the centrifuge test tube (highlighted in Table \ref{Tab:evaluationModel}). Therefore, this object was chosen as the baseline for the subsequent experiment in this paper.

\begin{table}[h]
\caption{Evaluation results for trained model on collected dataset per object class}
\resizebox{0.48\textwidth}{!}{
\begin{tabular}{|l|c|c|c|c|c|c|c|}
\hline
\multicolumn{1}{|c|}{} & \multicolumn{4}{c|}{\textbf{\begin{tabular}[c]{@{}c@{}}Pixel metrics, \\ \%\end{tabular}}} & \multicolumn{3}{c|}{\textbf{\begin{tabular}[c]{@{}c@{}}Object metrics,\\ num out of 2000\end{tabular}}} \\ \cline{2-8} 
\multicolumn{1}{|c|}{\multirow{-2}{*}{\textbf{Object class}}} & \textbf{AP} & \textbf{AP$_{50}$} & \textbf{AP$_{75}$} & \textbf{AP$_{M}$} & \textbf{TP} & \textbf{FP} & \textbf{FN} \\ \hline
Cell scraper & 75.4 & 87.3 & 79.5 & 83.5 & 1784 & 35 & 216 \\ \hline
Micro test tube & 87.4 & 97.2 & 92.1 & 92.7 & 1927 & 23 & 73 \\ \hline
Needle holder & 86.8 & 96.9 & 90.2 & 91.1 & 1902 & 21 & 98 \\ \hline
Pasteur pipette & 76.3 & 88.1 & 84.9 & 83.7 & 1801 & 34 & 199 \\ \hline
Pipettor & 81.4 & 93.6 & 88.4 & 89.5 & 1847 & 28 & 153 \\ \hline
\rowcolor[HTML]{EFEFEF} 
\textbf{Centrifuge test tube} & {\textbf{88.2}} & {\textbf{97.5}} & {\textbf{93.2}} & {\textbf{93.8}} & {\textbf{1939}} & {\textbf{17}} & {\textbf{61}} \\ \hline
Vacuum test tube & 85.5 & 94.8 & 89.1 & 89.9 & 1893 & 29 & 107 \\ \hline
PCR tupfer & 77.0 & 86.3 & 82.1 & 84.6 & 1808 & 38 & 192 \\ \hline
\end{tabular}
}
\label{Tab:evaluationModel}
\vspace{-1em}
\end{table}


\subsection {R-CNN-based Object Recognition for Virtual Scene Augmentation}

After the trained neural network has shown sufficiently high performance in determining the bounding box and segmentation mask on the test images, we implemented an algorithm to estimate the positions of real objects. 

The RealSense camera produces two kinds of frames, a color one and a depth map. We used the color frame as input for the neural network to determine the bounding box and segmentation mask of the object, while, from the depth map, we got the distance to the object. In our system, we compare two ways of getting the distance to the object: the average value within the bounds of the object bounding boxes or the average value for the segmentation masks. Then the coordinates of the object are calculated using the built-in functions of the PyRealSense framework by projecting points of the object onto the plane and calculating coordinates along the axes. The object center's reconstruction point is the center of the obtained bounding box and segmentation mask. Thus, the class of a certain object, the center of its bounding box and segmentation mask, and the average distance to the object by the bounding box and the segmentation mask are transmitted to VE as information about the object.
For the better performance of the object position estimation in the VE, we use an alpha-filter to soft any fluctuation of the object position.
 
\section{User Study}

We conducted an experiment of robot teleoperation through two control modes, namely camera-based and VR-based teleoperation, to evaluate the convenience of using the GraspLook system and user ability to manipulate objects in the virtual environment. The first mode provided the operator with visual feedback from two cameras from the remote environment (on-gripper camera and isometric view camera). The VR environment provided a 360$^\circ$ view of the workspace by the user's head rotation. The performance of each task was characterized by execution time and the accuracy of the target object grasping. In addition, after completing the tasks using both teleoperation modes, we asked participants to respond to a 7-question survey using a seven-point Likert scale. The questionnaire was based on the NASA Task Load Index \cite{Nasa}. Besides, we added one additional question about the level of user involvement in the task. 
\subsection{Participants}
The experiment involved 8 subjects (5 males and 3 females). The average participant age was 24.2 (SD=1.5), with a range of 22–27. In total, three participants had never operated collaborative robots, three participants operated robotic arms several times, and two reported regular experience with collaborative robots. Six participants used VR only a few times, and two people answered that they used VR devices regularly.

\subsection{Experimental setup}
\begin{figure}[!t]
\centering
\includegraphics[width=0.95\linewidth]{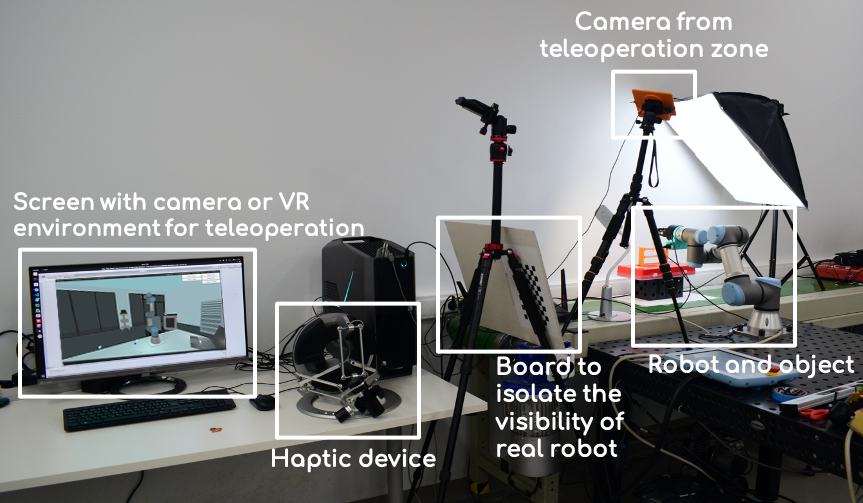}
\caption{Experimental setup for the experiment.}
\label{fig:teleoperationSetup}
\vspace{-1em}
\end{figure}

The first approach for teleoperation included only visual feedback from two cameras. The former (Logitech C930e) was used to display the teleoperation area with an isometric view of the robot and the object for the manipulation (test tube). The latter was mounted on the gripper (8 MP USB camera). For the first control mode, the operator was separated from the robot by an insulating board to exclude the view of the working area. The second system utilized the described in chapter \ref{VE} augmented visual feedback to the operator through a VE in the Unity Engine. The VR setup included HTC Vive Pro base stations, a head-mounted display (HMD), and a Vive controller to regulate the operator's point of view in VR. The virtual environment utilizes additional visual feedback from the remote environment through a video stream from the on-gripper camera. The experimental setup is shown in Fig. \ref{fig:teleoperationSetup}.

\begin{figure}[!h]
\centering
\includegraphics[width=0.8\linewidth]{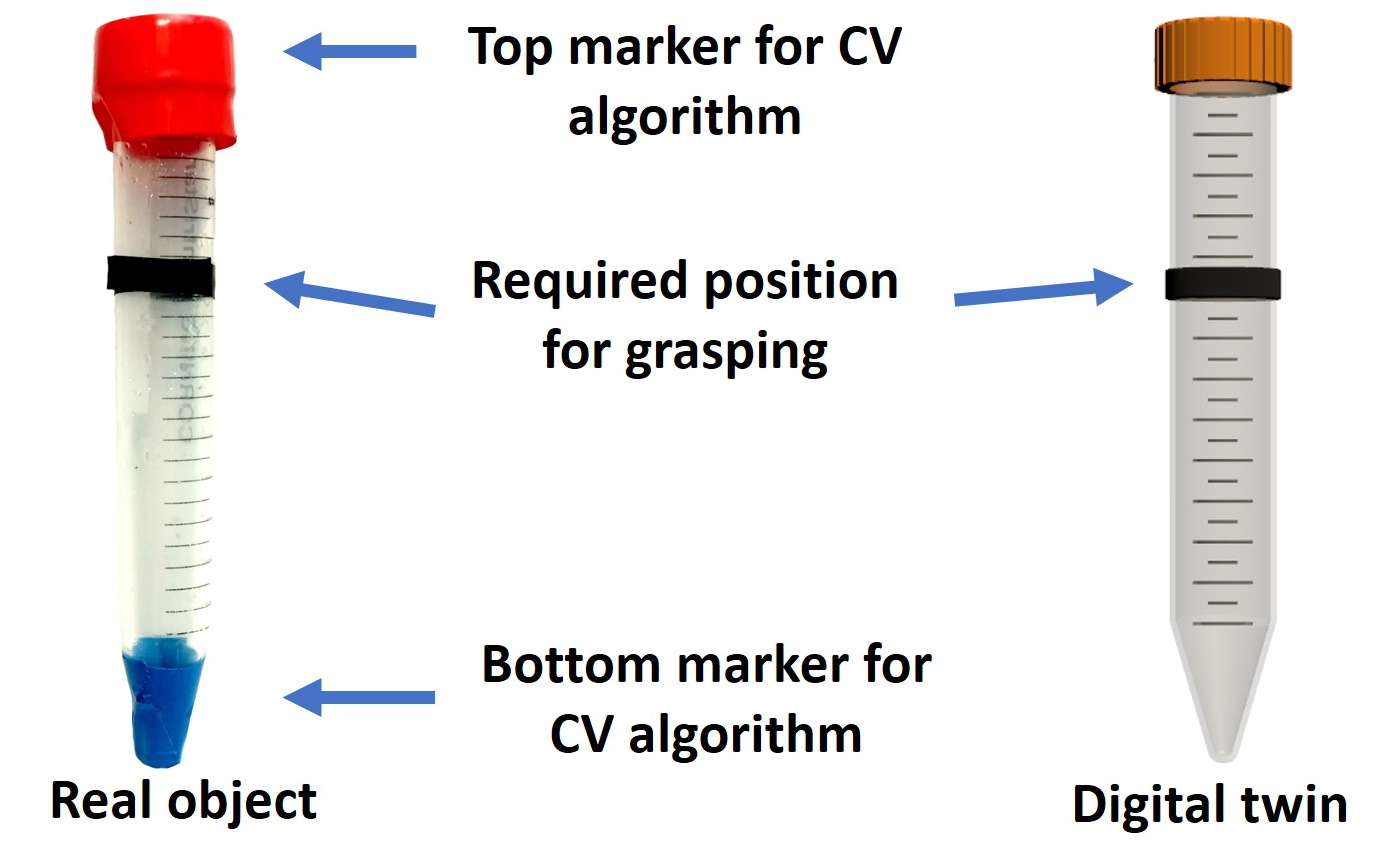}
\caption{The target object with a marked position for grasping.}
\label{fig:targetObject}
\vspace{-1em}
\end{figure}
\subsection{Experimental procedure}
The experimental task was to manipulate a test tube in a pick-and-place task. Each participant had three attempts for task completion. The participants were asked to grasp the target object at a marked position (Fig. \ref{fig:targetObject}). To improve the accuracy of target object pose estimation for VE augmentation, we developed an additional CV algorithm to determine the upper and lower parts of the target object by explicitly setting the color to these parts. Together with the bounding box definition, the color definition ensured that the object was recognized completely or with one of two possible halves (top or bottom). Since the distance to the center of the target object is known from each color markings, it reduced the error of determining the center of the object along the vertical axis.

At the beginning of the experiment, the EE of the robot was located at a distance of 47 $cm$ from the target object, and the camera was at the level of the object center. As the operator pressed the start button, the robot moved according to the manipulations with the Omega.7 haptic device. For each participant, we measured the following quantitative data for each control mode: task execution time, the accuracy of the grasping, measured as the error in the distance from the target point of grasping the object to the real point of grasping, the trajectory length of the robot, and the number of attempts for task completion.

\subsection{Experimental results}
After conducting the teleoperation experiment through both control modes, we compared the average trajectory length for each approach, the task execution time, and the accuracy of the object grasping (Table \ref{tab:teleoperationComparison}).
Since the experiment consisted of grasping the object and placing it in any hole of the tube rack, the length of the robotic gripper trajectory was measured from the moment the task was started to the moment of grasping the object. The experiment results showed that the average length of the trajectory that the robot passes before the object was the lowest for remote control of the robot in the VR mode ($0.43\ m$), which is also reflected in the average task execution time ($87.89\ s$ for the VR mode). Thus, developed VR interface allows operator to reduce the length of the robot trajectory to the target object by 23.2\% and reduce task completion time by 31.3\% compared to the camera-based teleoperation approach.
In terms of accuracy of grasping, the VR-based system also showed a minimum average error of $14.1\ mm$ when grasping the object. 


\begin{table*}[!t]
\centering
\caption{Comparison of Trajectory length, Time of Operation and Error at Grasping, and Mean Number of Attempts for 2 Teleoperation Approaches}
\resizebox{0.7\textheight}{!}{
\begin{tabular}{|c|c|c|c|c|c|c|c|c|c|c|c|c|c|}
\hline
 & \multicolumn{4}{c|}{\textbf{Trajectory length, m}}  & \multicolumn{4}{c|}{\textbf{Task execution time, s}} & \multicolumn{4}{c|}{\textbf{Error at grasping, mm}} &  \\ \cline{2-13} 
\multirow{-2}{*}{\textbf{Teleoperation mode}} & \textbf{min} & \textbf{max} & \textbf{mean} & \textbf{sd} & \textbf{min} & \textbf{max} & \textbf{mean} & \textbf{sd} & \textbf{min} & \textbf{max} & \textbf{mean} & \textbf{sd} & \multirow{-2}{*}{\textbf{Attempts}}\\ \hline
\textbf{Camera-based} & {\color[HTML]{000000} 0.16} & {\color[HTML]{000000} 1.82} & {\color[HTML]{000000} 0.56} & {\color[HTML]{000000} 0.52} & {\color[HTML]{000000} 48.2} & {\color[HTML]{000000} 193.23} & {\color[HTML]{000000} 127.95} & {\color[HTML]{000000} 50.15} & {\color[HTML]{000000} 7.4} & {\color[HTML]{000000} 29.5} & {\color[HTML]{000000} 15.9} & {\color[HTML]{000000} 7.6} & 1.4 \\ \hline
\textbf{VR-based} & {\color[HTML]{000000} 0.15} & {\color[HTML]{000000} 1.20} & {\color[HTML]{009901} \textbf{0.43}} & {\color[HTML]{000000} 0.35} & {\color[HTML]{000000} 40.25} & {\color[HTML]{000000} 147.55} & {\color[HTML]{009901} \textbf{87.89}} & {\color[HTML]{000000} 36.9} & {\color[HTML]{000000} 6.2} & {\color[HTML]{000000} 22.3} & {\color[HTML]{009901} \textbf{14.1}} & {\color[HTML]{000000} 5.8} & 1.4\\ \hline
\end{tabular}
}
\vspace{-1.5em}
\label{tab:teleoperationComparison}
\end{table*}

\begin{figure}[!t]
\centering
\includegraphics[width=0.45\textwidth]{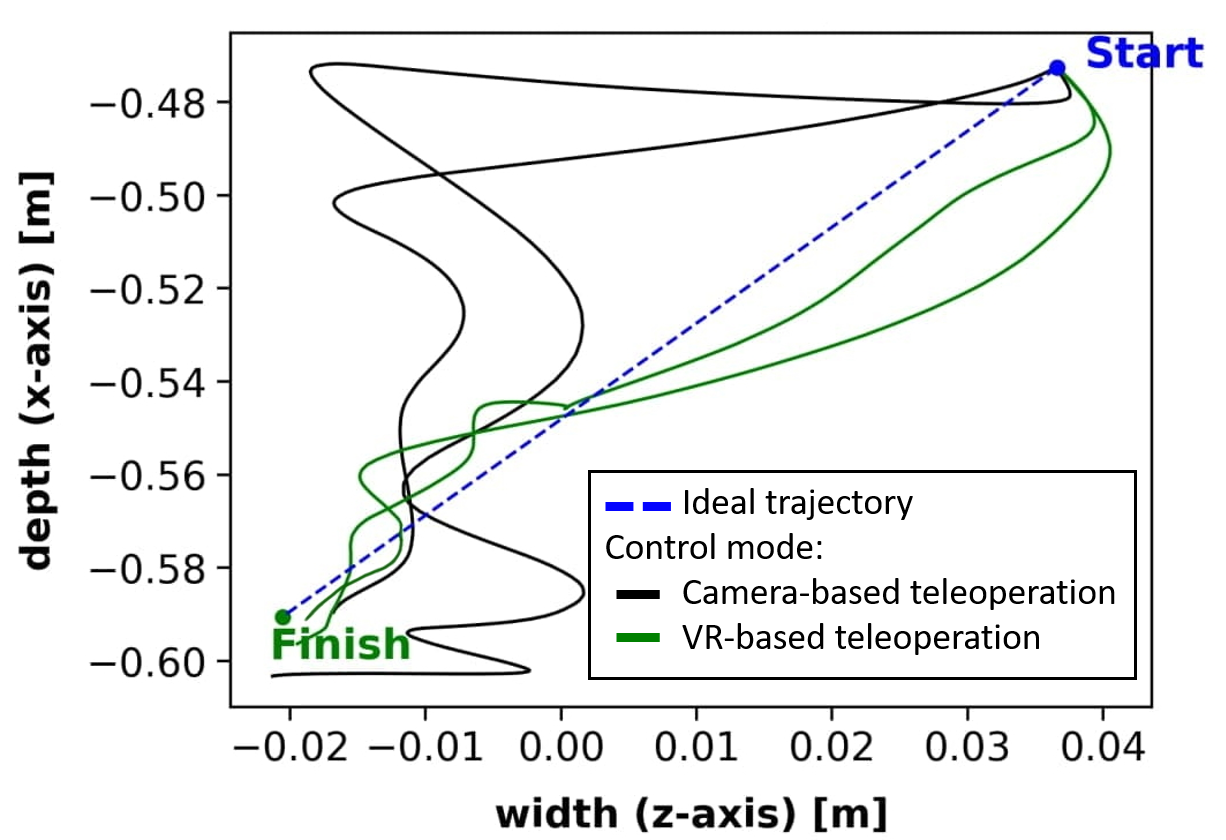}
\vspace{-1em}
\caption{Examples of the EE trajectories for 2 control modes. The blue dot represents the start position of the end effector and the green point is the position of the target object.}
\label{fig:trajectories}
\end{figure}

Fig. \ref{fig:trajectories} represents the example of the trajectories of EE moving from the initial position of the robot to the object grasping position for two participants. The experiment results for the trajectory length showed that for the camera-based control mode, the operator has to perform unnecessary movements since two views of the teleoperation zone are not enough.  
The trajectory of the robot movement for VR control mode has a more directional movement towards the target. 
An additional comparison of the robot trajectories is given in Table \ref{tab:comparisonTraj}. It shows a comparison of the trajectory lengths before grasping the object with the ideal trajectory of movement, which is a straight line between the initial point of the robotic gripper and the end point (the center of the grasped object). During the VR control mode, the robot trajectory increased by 186\% on average compared with the ideal trajectory for all participants.

\begin{table}[]
\caption{Comparison of Trajectory Length for Teleoperation Approaches Regarding Ideal Trajectory}
\resizebox{0.48\textwidth}{!}{
\begin{tabular}{|c|c|c|}
\hline
\textbf{Trajectory type} & \textbf{Mean length, m} &\textbf{\makecell{Increment regarding\\ ideal, \%}} \\ \hline
Ideal & 0.15 & 0\% \\ \hline
Camera-based teleoperation & 0.56 & 273\% \\ \hline
\textbf{VR-based teleoperation} & {\color[HTML]{009901} \textbf{0.43}} & {\color[HTML]{009901} \textbf{186\%}} \\ \hline
\end{tabular}
}
\vspace{-1.5em}
\label{tab:comparisonTraj}
\end{table}

Since participants have only three attempts to complete the task in the experiment, we compared the number of attempts with which participants coped with the pick-and-place task.
For both control modes, the mean number of attempts for task completion comprised 1.4 (Table \ref{tab:teleoperationComparison}). During a camera-based teleoperation, most participants could complete the task on the first trial. However, one participant took three attempts. During teleoperation using VR interface, most of the participants coped with the task on the first attempt, while the rest, getting used to the system, completed the task on the second attempt. For this system, no one had to use the third attempt.

\begin{figure}[!t]
\centering
\includegraphics[width=0.99\linewidth]{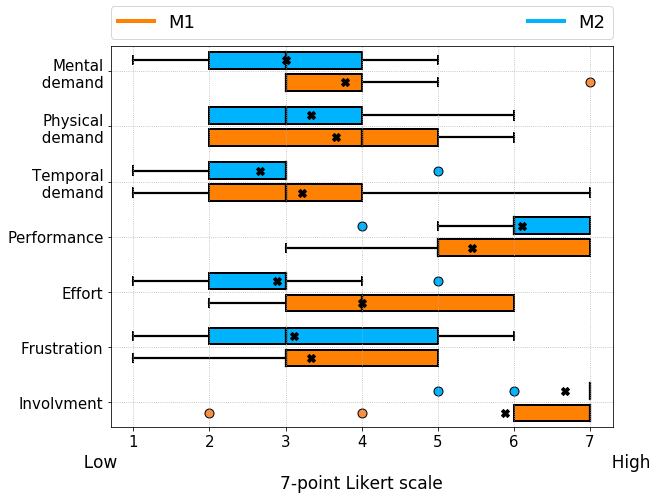}
\caption{Evaluation of the participant’s experience for both control methods in the form of a 7-point Likert scale (1 = Low, 7 = High). M1: camera-based teleoperation; M2: VR-based teleoperation. Crosses mark mean values.}
\label{fig:likert}
\vspace{-1em}
\end{figure}

Fig. \ref{fig:likert} shows the results of the user assessment of two control modes across all participants. According to the obtained results, VR-based interface increases involvement (by 11$\%$) and performance (by 10$\%$), whereas the frustration level during task completion and physical demand of participants were almost the same for both control modes. At the same time, participants noted that it is easier to control the system through the VR application. The teleoperation through VR interface did not require special mental costs and a lot of effort on the user's part, while the success of the completed task was felt better.

\section{Conclusions and Future Work}

We have proposed a novel telemanipulation system with R-CNN-based object recognition for virtual scene augmentation. Objects for manipulation detected using the developed CV system are displayed in the VE in real time, allowing the operator to perform precise manipulations.
The results of the user study revealed that the teleoperation through developed VR interface allowed users to operate the robot smoother, resulting in a shorter trajectory of the robot to the target object (by 23.2\%) and faster task completion (by 31.3\%) compared to the camera-based teleoperation approach. In addition, GraspLook system is less mentally demanding (by 11\%) and requires less effort (by 16\%) to robot manipulation than the camera-based teleoperation system while increasing user involvement by 11\%.

To further improve the system, we consider equipping the robotic gripper with tactile sensors to detect the object grasping and provide kinesthetic feedback to the user. 
In this paper, the experiments were carried out using a single object from the dataset - the centrifuge test tube, which has the highest detection accuracy value with proposed R-CNN. Besides, we are going to conduct experiments with each object class from the entire dataset, as well as R-CNN model evaluation in ablation study based on the dataset with real objects.
In addition, teleoperation in the environment with object occlusion will be experimentally evaluated in the future.

\section*{Acknowledgment}

The authors would like to thank Miguel Altamirano Cabrera, PhD student at Skolkovo Institute of Science and Technology, and Jonathan Tirado, PhD student at University of Southern Denmark, for their support to this project.

\bibliographystyle{IEEEtran}
\bibliography{bib}

\addtolength{\textheight}{-12cm} 


\end{document}